\documentclass[conference]{IEEEtran}
\IEEEoverridecommandlockouts
\usepackage{cite}
\usepackage{amsmath,amssymb,amsfonts}
\usepackage{algorithmic}
\usepackage{graphicx}
\usepackage{textcomp}
\usepackage{xcolor}
\usepackage{multirow}
\usepackage{multicol}
\usepackage{subcaption}
\def\BibTeX{{\rm B\kern-.05em{\sc i\kern-.025em b}\kern-.08em
    T\kern-.1667em\lower.7ex\hbox{E}\kern-.125emX}}
\begin{document}

\title{STRAP: Spatial-Temporal Risk-Attentive Vehicle Trajectory Prediction for Autonomous Driving\\
\thanks{This work is supported by the C2SMARTER, a Tier 1 U.S. Department of Transportation (USDOT) funded University Transportation Center (UTC) led by New York University funded by USDOT.}
}

\author{Xinyi Ning$^1$, Zilin Bian$^{1}$\textsuperscript{,*}, Dachuan Zuo$^1$, Semiha Ergan$^{1}$\textsuperscript{,*}
\thanks{$^1$ Tandon School of Engineering, New York University, NY, USA, 10012. }%

\thanks{* Corresponding Authors emails: {\tt\small \{zb536, semiha\}@nyu.edu}}%
}

\maketitle

\begin{abstract}
Accurate vehicle trajectory prediction is essential for ensuring safety and efficiency in fully autonomous driving systems. While existing methods primarily focus on modeling observed motion patterns and interactions with other vehicles, they often neglect the potential risks posed by the uncertain or aggressive behaviors of surrounding vehicles. In this paper, we propose a novel spatial-temporal risk-attentive trajectory prediction framework that incorporates a risk potential field to assess perceived risks arising from behaviors of nearby vehicles. The framework leverages a spatial-temporal encoder and a risk-attentive feature fusion decoder to embed the risk potential field into the extracted spatial-temporal feature representations for trajectory prediction. A risk-scaled loss function is further designed to improve the prediction accuracy of high-risk scenarios, such as short relative spacing. Experiments on the widely used NGSIM and HighD datasets demonstrate that our method reduces average prediction errors by 4.8\% and 31.2\% respectively compared to state-of-the-art approaches, especially in high-risk scenarios. The proposed framework provides interpretable, risk-aware predictions, contributing to more robust decision-making for autonomous driving systems.
\end{abstract}


\section{Introduction}
As autonomous vehicles (AVs) become more integrated into everyday transportation systems, their safety remains a significant concern\cite{b1}. One of the key factors in achieving safe autonomous driving is accurate vehicle trajectory prediction, which aims to infer a vehicle’s future movements based on its historical positions. A precise forecast of the movements of both the autonomous vehicle and surrounding traffic can help the system in the decision-making process about future routes, thus reducing the likelihood of accidents and ensuring safety and efficiency. 

To ensure safety in mixed traffic environments where AVs and human-driven vehicles coexist, it is essential to account for the potential risks posed by surrounding vehicles\cite{b30}. Researches have found that human drivers subconsciously evaluate potential dangers by considering the proximity, velocity, and motion directions of surrounding vehicles, implicitly linking perceived risks to their driving decisions\cite{b29}. Current AV trajectory prediction frameworks typically lack this nuanced, behavior-informed risk assessment capability, potentially compromising prediction accuracy. Therefore, incorporating a risk assessment model inspired by human driver's risk perception into trajectory prediction models may enhance predictive accuracy and offer deeper insights into AV safety, decision-making, and compatibility with human-driven vehicles.


\begin{figure}[htbp]
\centering
\begin{subfigure}[b]{0.42\textwidth}
    \includegraphics[width=\textwidth]{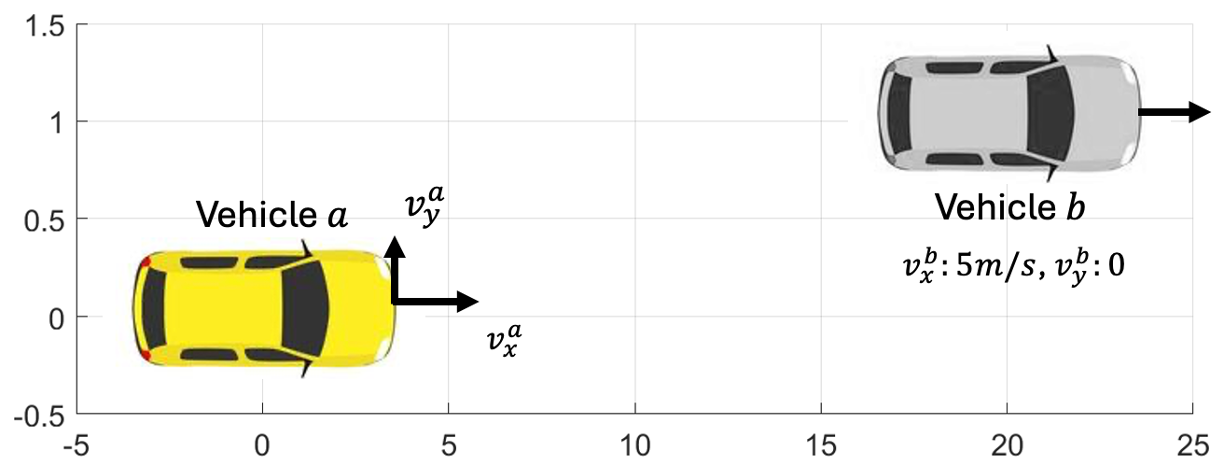}
    \caption{Relative distance and speed between vehicle \(a\) and \(b\)}
    \label{fig:sub1}
\end{subfigure}
\hfill
\begin{subfigure}[b]{0.21\textwidth}
    \includegraphics[width=\textwidth]{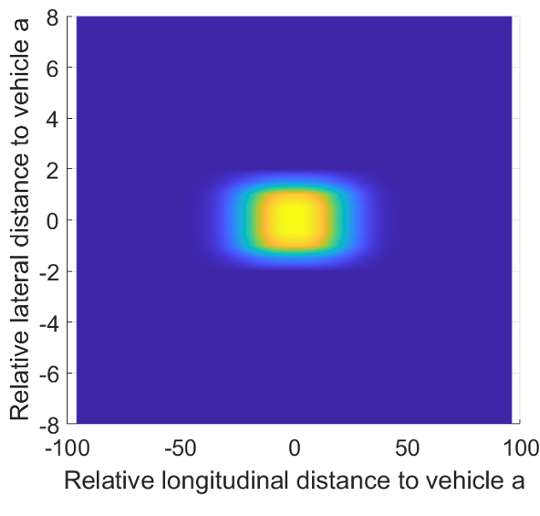}
    \caption{Spatial proximity risk perceived by vehicle \(a\)}
    \label{fig:sub2}
\end{subfigure}
\begin{subfigure}[b]{0.27\textwidth}
    \includegraphics[width=\textwidth]{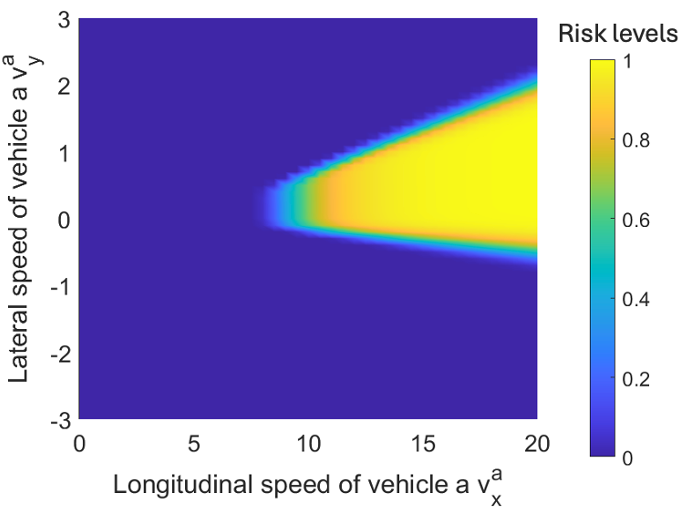}
    \caption{Temporal proximity risk perceived by vehicle \(a\)}
    \label{fig:sub3}
\end{subfigure}
\caption{An example of a risk field measuring the spatial proximity of surrounding objects and the temporal proximity to a collision}
\label{example_fig}
\end{figure}

In real-world driving, risk is not a fixed value but a dynamic distribution that evolves continuously over both time and space. Prior research has attempted to incorporate risk into trajectory prediction using surrogate safety measures, such as Time-To-Collision (TTC)\cite{b28}, which quantifies the remaining time before a collision would occur if both vehicles maintain their current speeds. However, these metrics are inherently point-based and static, unable to adequately capture the continuous and dynamic spatial-temporal interactions among vehicles. As illustrated in Fig.~\ref{example_fig}, a risk field integrating both spatial and temporal proximities more effectively reflects the evolving nature of threats, thereby assessing safety more holistically. While previous studies\cite{b6} have explored the use of risk fields for trajectory prediction, the risk fields employed have predominantly considered only spatial proximity, neglecting crucial temporal factors like the time remaining to a potential collision, which significantly influence human risk perception. Therefore, developing a trajectory prediction framework that comprehensively quantifies both spatial and temporal dimensions of risk is essential for accurately modeling vehicle interactions and enhancing the reliability and precision of trajectory prediction.


Building on these insights, we propose a safety-oriented trajectory prediction model that can assess both the spatial proximity risk between vehicles and the probability of collision. The contributions of this study are as follows:

1) We present STRAP, a spatial-temporal risk-attentive trajectory prediction framework which fuses spatial-temporal dynamics and risk potential mapping into trajectory prediction. STRAP leverages multi-head attention mechanisms, specifically implementing a spatial-temporal encoder and a risk-attentive feature fusion module that computes the risk distribution across future intention endpoints and integrates it with spatial-temporal features for trajectory prediction.

2) A new loss function, scaled by spatial and temporal proximity risk levels, is presented to prioritize the prediction accuracy of high-risk scenarios during training, encouraging the model to focus on safety-critical scenarios where prediction errors have severe consequences. 

The proposed framework is evaluated using two public datasets, Next Generation Simulation (NGSIM) and Highway Drone (HighD).  

This paper is structured as follows. In Section II, we give an overview of related past work. Section III presents STRAP, the risk-attentive trajectory prediction framework. Section IV evaluates the framework using real-world datasets. Section V summarizes the conclusions.

\section{Related Works}

\subsection{Deep Learning Models for Trajectory Prediction}

Compared with traditional dynamics-based models, deep learning methods have demonstrated strong capabilities in capturing complex vehicle behavior patterns and multi-agent interactions. Many studies reformulate trajectory prediction as a time series forecasting task, using deep learning models to learn the spatial and temporal dependencies in vehicle trajectories. Convolutional Neural Networks (CNNs)\cite{b7} and Graph Neural Networks (GNNs) \cite{b9} have been widely adopted for spatial feature extraction, while Recurrent Neural Networks (RNNs)\cite{b10}, Long-Short Term Memory (LSTMs)\cite{b11,b12}, Gated Recurrent Units (GRUs)\cite{b13} are commonly used for temporal modeling. Transformers, with their self-attention mechanism, can effectively capture both long-term and short-term dependencies by processing sequences holistically, and have been increasingly adopted in trajectory prediction\cite{b2,b15}. Moreover, generative approaches such as diffusion models \cite{b17} and Generative Adversarial Networks (GANs)\cite{b18} have also been introduced to learn the probability distribution of future trajectories.

In addition to pattern-based data-driven methods, physics-based models and goal-based models have also been explored in many studies\cite{b19,b20}. Physics-based models combine deep learning with physics-informed principles to produce more realistic and robust trajectory predictions. For example, Liao et al. (2024) proposed a wavelet reconstruction network for handling missing observations, incorporating a kinematic bicycle model to ensure consistency with real-world vehicle dynamics\cite{b19}. Sheng et al. (2024) introduced a kinematics-aware multi-graph attention network that integrates bicycle and unicycle kinematic models into a deep graph-attention framework\cite{b20}. 

Goal-based methods typically begin by predicting a set of possible destination candidates for the target vehicle along with their probabilities, and then generating continuous trajectories based on each candidate. For example, Zhao et al. (2021) proposed a target-driven trajectory prediction framework that follows this approach and ultimately ranks and selects a compact set of high-probability trajectories\cite{b21}. Similarly, Gan et al. (2024) introduced a two-stage model that first predicts the vehicle's goal and then generates the corresponding trajectories by integrating neural networks with a physics-based social force model\cite{b22}. Our method is inspired by goal-based trajectory prediction, where future intention endpoints are estimated, and leverages a combination of Transformer and LSTM architectures for trajectory prediction.


\subsection{Risk Potential Field Models}
Risk assessment is essential for AVs to drive safely and anticipate potential hazards. To capture the continuous variation of risk in space and quantify the likelihood of collisions, various risk potential field models have been proposed. Kolekar et al. (2020) proposed a Driver’s Risk Field (DRF) that calculates a driver’s subjective probability of occupying a given position and estimates perceived crash risk by multiplying this probability by the potential consequence\cite{b23}. Wang et al. (2022) introduced a prediction‐based probabilistic driving risk field for multi‐lane highways, incorporating lane-change intention, future trajectories, collision probability, and severity into the risk estimation\cite{b24}. Zuo et al. (2025) developed a composite safety potential field consisting of a subjective field, which quantifies human-like perception of proximity risk, and an objective field, which captures imminent collision risk\cite{b25}. Although there have been various risk potential field models, existing trajectory prediction methods have not fully leveraged them to enhance their predictive accuracy, particularly in high-risk scenarios.


\section{Method}
\subsection{Problem Formulation}\label{AA}
The goal of trajectory prediction is to learn a mapping from historical positions of the target vehicle and its surrounding agents to the target’s future trajectory. The input sequence is denoted as \(\mathbf{S}_{T-T_h:T}^{0:N_v}=(s_{T-T_h:T}^{i})\), representing the history states over past \(T_h\) timesteps for the target vehicle \((i=0)\) and its \(N_v\) neighboring vehicles \((1\leq i\leq N_v)\). Each state \(s_{t}^{i}\) includes the 2D coordinates relative to the target vehicle, velocity, acceleration, vehicle dimensions, types, and lane ID at timestep \(t\), for both the target and neighboring vehicles. To improve the model's risk awareness, we adopt a risk-aware neighborhood selection strategy. Instead of selecting nearby vehicles purely based on distance, we calculate the potential risk between the target vehicle and all other vehicles in the scene using the potential field model described in Section~\ref{sec:risk}, and prioritize those associated with higher risk. This risk-aware selection strategy enables the model to focus on safety-critical interactions between the target and neighboring vehicles. Furthermore, we augment the vehicle state inputs with additional risk features, denoted as \(\mathbf{R}_{T-T_h:T}^{0:N_v}=(r_{T-T_h:T}^{i})\), where each \(r_{T-T_h:T}^{i}\) represents the subjective and objective risks perceived by vehicle \(i\). 

To address the variability and uncertainty in vehicle movements, we apply distribution learning to the output of the framework. Specifically, the framework outputs a distribution \(\mathbf{Y}_{T+1:T+T_f}^0\) representing the predicted positions of the target vehicle from time \(T+1\) to time  \(T+T_f\). At each future timestep \(t\), the position \(\mathbf{Y}_{t}^0\) is modeled as a bivariate Gaussian distribution with mean \(\hat{\mu}^0_t=(\hat{\mu}_x,\hat{\mu}_y)_{t}^0\), variances  \(\hat{\sigma}^0_t=(\hat{\sigma}_x,\hat{\sigma}_y)_{t}^0\) and a correlation coefficient \(\hat{\rho}_{t}^0\)\cite{b2}. This distribution is de-facto a probability density function capturing the likelihood of various potential positions of the vehicle at each future timestep.

\subsection{Risk Potential Field}
\label{sec:risk}
We adopt the safety potential field model for risk assessment from \cite{b24}. The driving risk consists of two parts: subjective risk (S-field) and objective risk (O-field). 


S-field uses a generalized Gaussian distribution model to quantify the driver's perceived proximity risk. The closer a surrounding vehicle intrudes the safety space of the target vehicle, the greater the risk and the greater pressure is perceived by the driver. The two-dimensional S-field risk \(r_{ij}^s\) perceived by vehicle \(i\) due to the proximity of a surrounding vehicle \(j\) is formulated in \eqref{eq1}:
\begin{equation}
r_{ij}^s=\exp\left(-\left|\frac{\Delta x_{ij}}{\gamma_x}\right|^{\alpha_x}-\left|\frac{\Delta y_{ij}}{\gamma_y}\right|^{\alpha_y}\right)\label{eq1}
\end{equation}
where \(\Delta x_{ij}\) and \(\Delta y_{ij}\) represent the longitudinal and lateral distances between the two vehicles. \(\gamma_x>1\), \(\gamma_y>1\) are the longitudinal and lateral scaling factors, and \(\alpha_x\geq2\), \(\alpha_y\geq2\) are the corresponding shape factors.

O-field estimates the probability of collision between two vehicles by predicting the potential collision point in the future and calculating the time proximity to that collision. The O-field risk \(r_{ij}^o\) perceived by vehicle \(i\) from vehicle \(j\) is defined in \eqref{eq2}:
\begin{equation}
r_{ij}^o=\exp\left[-\left(\frac{\hat{d}_{m,ij}}{d^*}\right)^{\beta_1}\right]\exp\left[-\left(\frac{\hat{t}_{m,ij}}{t^*}\right)^{\beta_2}\right]\label{eq2}
\end{equation}
where \(\hat{d}_{m,ij}\) is the predicted future minimum distance between the two vehicles and \(\hat{t}_{m,ij}\) is the time frame when the gap between the vehicles stops narrowing.  \(d^*\), \(t^*\) are scaling factors, and \(\beta_1\), \(\beta_2\) are the shape factors.

\subsection{Basic Architecture of STRAP}
\begin{figure}[htbp]
\centerline{\includegraphics[width=0.5\textwidth]{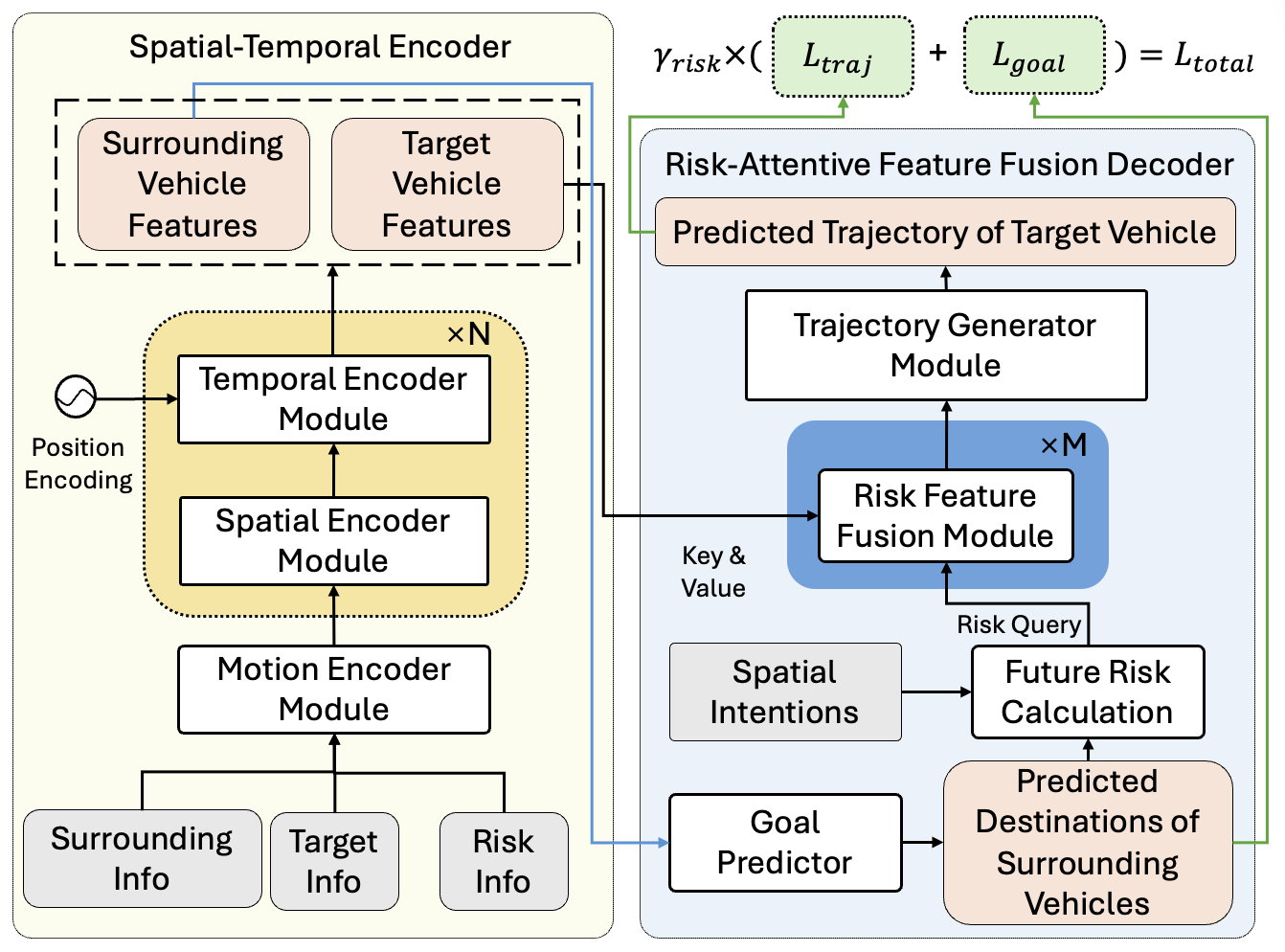}}
\caption{Model architecture of STRAP}
\label{fig1}
\end{figure}

The overall architecture of STRAP is illustrated in Fig.~\ref{fig1}. It adopts a standard encoder-decoder architecture. The encoder learns temporal and spatial features from the observed trajectories and associated risks, capturing motion dynamics and contextual interactions. Based on the encoded features of surrounding vehicles, the decoder first predicts their future goal locations and then estimates the corresponding risk levels for each potential intention mode of the target vehicle. These risk estimates are subsequently fused with the encoded features of the target vehicle to generate its trajectory predictions over the forecasting horizon. The entire framework is trained end-to-end using a combination of goal prediction loss and trajectory prediction loss, weighted by a risk-aware scaling factor.

\subsection{Spatial-Temporal Encoder}
The spatial-temporal encoder consists of three modules: a motion encoder, a spatial encoder, and a temporal encoder. The motion encoder employs a fully connected layer to transform the input vehicle states and risks into embedding representations, with an Exponential Linear Unit (ELU) as the activation function. These embeddings are then processed by an LSTM layer to capture the temporal correlations and motion patterns. 

The spatial encoder first computes spatial attention using a multi-head self-attention layer, capturing the relative importance and interactions among all vehicles. A Gated Linear Unit (GLU) is then applied to further extract the social dependencies among the vehicles. The overall spatial encoder is defined as:
\begin{equation}
\tilde{H}_s=\text{MultiHeadAttn}(q=H_m,k=H_m,v=H_m)\label{eq3}
\end{equation}
\begin{equation}
H_s=\text{LayerNorm}(\text{GLU}(\tilde{H}_t)+\tilde{H}_t)\label{eq4}
\end{equation}
where MultiHeadAttn is the multi-head attention layer and LayerNorm denotes Layer Normalization. \(H_m\in\mathbb{R}^{T_h\times (N_v+1)\times D}\) is the output of the motion encoder, \(D\) is the feature dimension, \(T_h\) is the number of historical timesteps, and \(H_s\in\mathbb{R}^{T_h\times (N_v+1)\times D}\) is the output of the spatial encoder.

The temporal encoder calculates the temporal dependency across different timestamps for each vehicle. Similar to the spatial encoder, a multi-head attention layer is used following a GLU layer, formulated as:
\begin{equation}
\bar{H}_s=H_s+PE\label{eq5}
\end{equation}
\begin{equation}
\tilde{H}_t=\text{MultiHeadAttn}(q=\bar{H}_s,k=\bar{H}_s,v=\bar{H}_s)\label{eq6}
\end{equation}
\begin{equation}
C=\text{LayerNorm}(\text{GLU}(\tilde{H}_t)+\tilde{H}_t)\label{eq7}
\end{equation}
where \(C\in\mathbb{R}^{(N_v+1)\times T_h\times D}\) is the output of the temporal encoder, and also the final output of the spatial-temporal encoder. \(D\) is the feature dimension and PE is the standard sinusoidal positional encoding. 

The spatial and temporal encoder layers can be stacked \(N\) times to capture more hierarchical and abstract representations of the input features, where \(N\) represents the number of encoding layers. 

\subsection{Risk-Attentive Feature Fusion Decoder}
\begin{figure}[htbp]
\centerline{\includegraphics[width=0.5\textwidth]{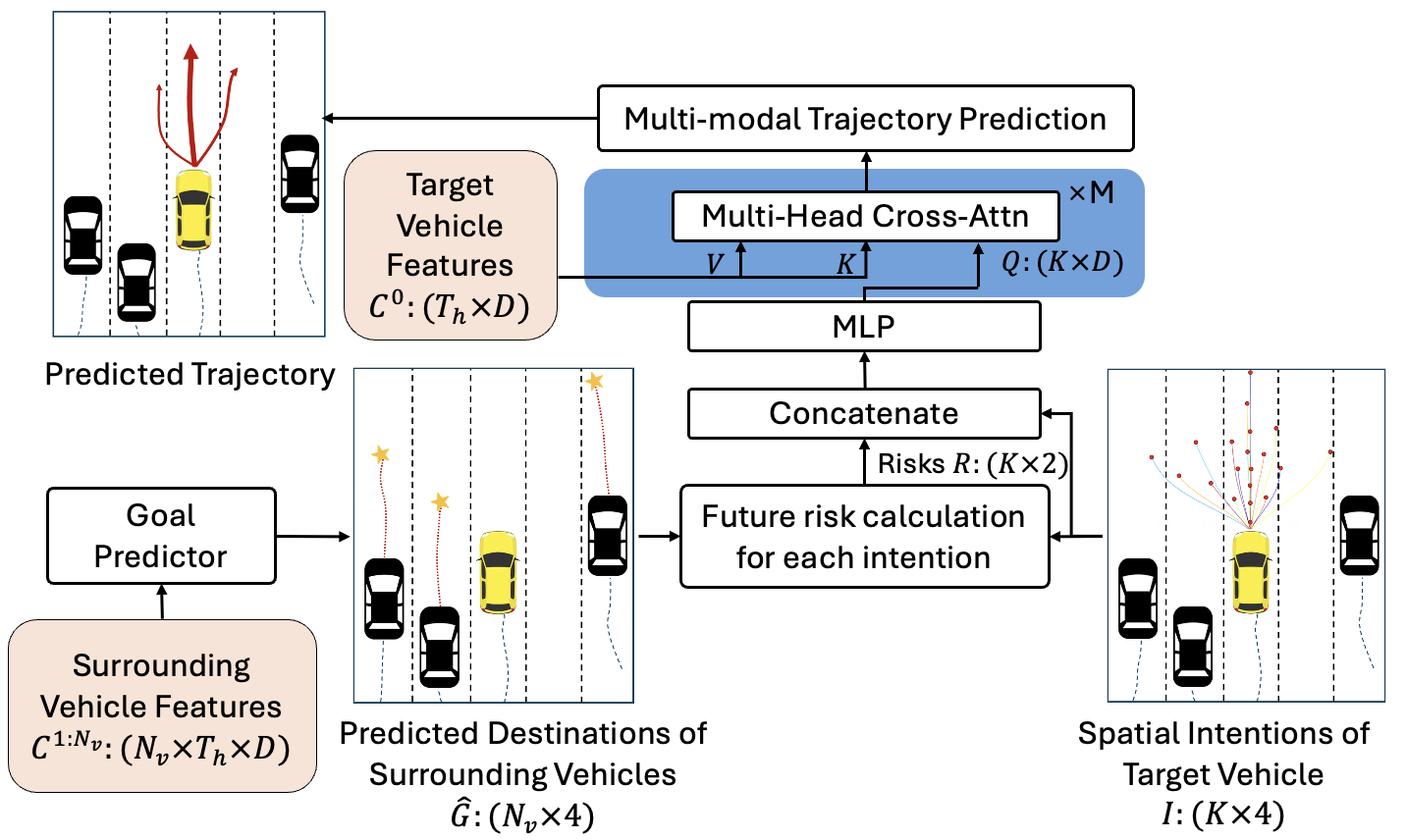}}
\caption{Architecture of the risk-attentive feature fusion decoder}
\label{fig2}
\end{figure}

The architecture of the risk-attentive feature fusion decoder is depicted in Fig.~\ref{fig2}. It integrates the encoded features with the prospective risk associated with each spatial intention mode of the target vehicle to enhance trajectory prediction. Each spatial intention mode represents a possible destination the target vehicle may reach within the forecasting horizon, characterized by its longitudinal and lateral position as well as velocity at the final predicted timestep. Inspired by the previous trajectory prediction model MTR\cite{b15}, we apply k-means clustering on the final positions of the ground-truth (GT) trajectories to derive \(K\) different intention modes, denoted as \(\mathbf{I}\in\mathbb{R}^{K\times4}\).

Furthermore, we include a goal predictor module to estimate the future destinations of surrounding vehicles based on their encoded features \(C^{1:N_v}\in\mathbb{R}^{N_v\times T_h\times D}\). This module first flattens the \(T_h\times D\) feature matrix of each vehicle into a single vector. It is then passed through a multi-layer perceptron (MLP), which processes the flattened features to predict the final position and velocity goals \(\hat{G}^{1:N_v}\in\mathbb{R}^{N_v \times 4}\) for each vehicle at the end of the prediction horizon, formulated as:
\begin{equation}
\hat{G}^{1:N_v}=\text{MLP}(C^{1:N_v})\label{eqgoal}
\end{equation}


Given the predicted goals of surrounding vehicles \(\hat{G}^{1:N_v}\), we compute the subjective risk \(R^s=\sum_{j=1}^{N_v}r^s_{0j}\) and objective risk \(R^o=\sum_{j=1}^{N_v}r^o_{0j}\) for each intention mode using the potential field model detailed in Section~\ref{sec:risk}. Here, \(r^s_{0j}\) and \(r^o_{0j}\) denote the subjective and objective risks between the target vehicle and the \(j\)-th surrounding vehicle, respectively. These risks quantify the perceived danger and estimated collision probability when the target vehicle follows a given intention mode and reaches the corresponding destination. By computing the risk values \(R=(R^s,R^o)\) for all \(K\) intention modes and concatenating them with the spatial intention vectors \(\mathbf{I}\in\mathbb{R}^{K\times4}\), a discrete spatial field of future potential risk distribution is obtained.

We further embed the risk field into high-dimensional representations \(H_q^0\in\mathbb{R}^{K\times D}\) using an MLP layer, where \(D\) denotes the feature dimension, to facilitate subsequent cross-attention computations. These embeddings \(H_q^0\) are used as risk field queries, while the encoded features of the target vehicle \(C^0\) serve as the values and keys. Cross-attention is computed as follows:
\begin{equation}
H_q^0=\text{MLP}(\left[\mathbf{R},\mathbf{I}\right])\label{eq9}
\end{equation}
\begin{equation}
H_q^j=\text{MultiHeadCrAttn}(q=H_q^{j-1},k=C^0,v=C^0)\label{eq10}
\end{equation}
where MultiHeadCrAttn denotes multi-head cross attention, \(\mathbf{R}\in\mathbb{R}^{K\times2}\) represents the subjective and objective risk field, \(C_0\in\mathbb{R}^{T_f\times D}\) is the encoded features of the target vehicle and \(H_q^j\in\mathbb{R}^{K\times D}\) is the risk field query at the j-th attention layer.

The final layer of the risk field query \(H_q^M\) is passed to the trajectory generator module, which employs an LSTM followed by an MLP to generate a probability distribution \(\hat{\mathbf{Y}}_{T+1:T+T_f}^0\in\mathbb{R}^{T_f\times5}\) for the future trajectory of the target vehicle.

\subsection{Risk-Scaled Loss Function}
 The loss function consists of two parts: the goal prediction loss \(\mathcal{L}_{goal}\) of the surrounding vehicles and the trajectory prediction loss \(\mathcal{L}_{traj}\) of the target vehicle. We use Mean Square Error (MSE) for \(\mathcal{L}_{goal}\), and both MSE and Negative Log-Likelihood (NLL) loss for \(\mathcal{L}_{traj}\), defined as:
 \begin{equation}
\mathcal{L}_{goal}=||\hat{G}^i-G^{i,gt}||^2\label{eq12}
\end{equation}
\begin{equation}
\mathcal{L}_{traj}=\frac{1}{T_f}\sum_{t=T+1}^{T+T_f}\left[||\hat{\mu}_t^i-{\mu}^{i,gt}_t||^2-log(\mathcal{N}(\mathbf{\mu}_t^{i,gt}|\hat{\mu}^{i}_t,\hat{\sigma}_t^i,\hat{\rho}_t^i))\right]\label{eq13}
\end{equation}
where \(\hat{G}^i\) and \(G^{i,gt}\) are the predicted and ground truth endpoints of vehicle \(i\), and \({\mu}^{i,gt}_t=({\mu}_x,{\mu_y})^{i,gt}_t\) is the ground truth coordinate of vehicle \(i\) at time \(t\).

 To enhance the model’s focus on safety-critical scenarios, we introduce a risk scaling factor \(\gamma_{risk}\), which is based on the sum of the subjective and objective risks \(R^s\) and \(R^o\) perceived by the target vehicle. This factor increases the contribution of high-risk samples to the loss: for the same MSE or NLL value, a higher risk level results in a larger total loss. By assigning greater weight to such cases, the model is encouraged to learn more discriminative features in situations where the consequences of prediction errors are potentially severe, thereby improving the prediction accuracy of safety-critical scenarios. The overall loss \(\mathcal{L}_{total}\) is defined as:
\begin{equation}
\gamma_{risk}=\max\left[\exp(R^s+R^o)-\beta,1\right]\label{eq10}
\end{equation}
\begin{equation}
\mathcal{L}_{total}=\gamma_{risk}(\mathcal{L}_{goal}+\mathcal{L}_{traj})\label{eq11}
\end{equation}
where \(\beta\) is a pre-defined bias term.

\section{Experiments}
In this section, we discuss the experimental setups and results of STRAP's performance and compare them with state-of-the-art algorithms.

\subsection{Experimental Setup}
\paragraph{Dataset} The experiment employs two datasets: NGSIM\cite{b26} and HighD\cite{b27}. These datasets are derived from real-world highway settings, including detailed vehicle trajectory data such as 2D positions, velocities, accelerations, vehicle dimensions, and lane information over time. We segment the data into 8-second intervals, using the first 3 seconds \((t_h=3)\) as the input trajectory history and the following 5 seconds \((t_f=5)\) as the ground truth for future prediction. Each vehicle in a scene is treated as a target vehicle, resulting in 7.7M trajectories in the NGSIM dataset and 17.7M trajectories in the HighD dataset. The data is divided into training, validation, and testing sets with a split ratio of 0.7, 0.1, and 0.2, respectively.


\paragraph{Implementation Details} The experiment was run on a machine with NVIDIA GeForce RTX 4090 GPU and an Intel Core i9-14900KF CPU. We consider the vehicles with either S-field risk or O-field risk greater than 0.005 as interacting with the target vehicle and select them as its neighborhood vehicles. On average, each vehicle has 6 neighboring vehicles, and the maximum number of neighborhood vehicles is set to \(N_v=15\). The hidden feature dimension of the encoder and decoder is \(D=64\). The number of spatial intention modes in k-means clustering is \(K=100\). The model has 3 spatial and temporal encoder layers and 2 risk feature fusion decoder layers. It is trained in an end-to-end manner for 12 epochs with the Adam optimizer and a batch size of 128. The initial learning rate was 0.0005 and was decayed by a factor of 0.6 during each epoch.

\paragraph{Evaluation Metrics} The results have been evaluated using Root Mean Squared Error (RMSE). 


\paragraph{Comparison of STRAP with Other Models} We compare STRAP with four benchmarks, representing a diverse range of trajectory prediction approaches, including a traditional physics-based method, LSTM-based models, and a transformer-based model. 
\begin{itemize}
    \item CV: A physics-based model that assumes that the target vehicle moves at a constant velocity.
    \item Social-LSTM (S-LSTM)\cite{b11}: An LSTM-based model that introduces a social pooling layer to aggregate the features of different vehicles.
    \item Convolutional Social-LSTM (CS-LSTM)\cite{b12}: A LSTM-based model that uses convolutional social pooling to capture interactions.
    \item STDAN\cite{b2}: A transformer-based model that considers driver's intentions.
    \item STRAP-B: Our model using basic loss function of the sum of \(\mathcal{L}_{goal}\) and \(\mathcal{L}_{traj}\).
    \item STRAP-R: Our model using the risk-scaled loss function.
\end{itemize}

\subsection{Results}

Table~\ref{tab1} compares the prediction errors of STRAP to those of the baseline models. Overall, STRAP outperforms all baselines across every prediction horizon. On the NGSIM dataset, STRAP reduces RMSE by 14.0\%, 6.9\%, and 5.3\% for 1s, 2s, and 3s short-term predictions, respectively, compared to the best baseline (STDAN), and by 3.5\% and 3.8\% for 4s and 5s long-term predictions. On the HighD dataset, because of its high precision, the improvements are more significant with RMSE reduced by 33.3\%, 66.7\%, 43.8\%, 29.7\%, and 25.4\% over STDAN across the 1s to 5s prediction horizons. 


\begin{table}[htbp]
\caption{Prediction RMSE (m) of STRAP and the benchmark models}
\begin{center}
\begin{tabular}{|c|c|c|c|c|c|c|c|}
\hline
\multirow{2}{*}{Dataset}&\multirow{2}{*}{Methods} &\multicolumn{6}{|c|}{Prediction Horizon (s)}\\
\cline{3-8} 
&  & 1 & 2 & 3 & 4 & 5 & Average\\
\hline
\multirow{6}{*}{NGSIM} & CV & 0.59 & 1.59 &2.87  &4.44 & 6.25 & 3.15\\
 & S-LSTM  & 0.65 & 1.31 & 2.16 & 3.25 &  4.55  & 2.38\\
 & CS-LSTM & 0.57 & 1.26 & 2.11 & 3.18 &  4.50  & 2.32\\
 & STDAN  & 0.43 & 1.01 & 1.69 & 2.56 & 3.65  & 1.87\\
 & STRAP-B & 0.38 & \textbf{0.94} & \textbf{1.60} & 2.47 & \textbf{3.51} & \textbf{1.78}\\
 &  STRAP-R &\textbf{0.37} & 0.94 & 1.61 & \textbf{2.47} & 3.52  & 1.78\\
\hline
\multirow{6}{*}{HighD} & CV   & 0.18 & 0.68 & 1.49 & 2.55 &  3.86  & 1.75\\
 & S-LSTM  & 0.22 & 0.62 & 1.27 & 2.15 & 3.41  & 1.53\\
 & CS-LSTM  & 0.22 & 0.61 & 1.24 & 2.10 & 3.27  & 1.49\\
 & STDAN & 0.06 & 0.15 & 0.32 & 0.64 & 1.22 & 0.48 \\
 & STRAP-B & 0.04 & 0.05 & 0.18 & 0.46 & 0.93  & 0.33\\
 & STRAP-R & \textbf{0.04} & \textbf{0.05} & \textbf{0.18} & \textbf{0.45} & \textbf{0.91} & \textbf{0.33} \\
 \hline
\end{tabular}
\label{tab1}
\end{center}
\end{table}


To further evaluate the effect of the risk-scaled loss function, Table~\ref{tab2} compares the average RMSE of our model when trained with the basic loss function (STRAP-B) versus the risk-scaled loss function (STRAP-R) across scenarios in NGSIM at different risk levels. In general, prediction accuracy is lower when collision probability is higher and vehicle behavior becomes less predictable. However, by incorporating the risk scaling factor, STRAP-R successfully captures high-risk trajectories and learns their discriminative features. Although STRAP-B and STRAP-R achieve similar average RMSE overall, as shown in Table~\ref{tab1}, STRAP-R shows a clear advantage under higher-risk scenarios in Table~\ref{tab2}, demonstrating its ability to predict more realistic trajectories that represent human driving behaviors. 





\begin{table}[htbp]
\caption{Average RMSE (m) in NGSIM on scenarios at different risk levels over 5s prediction}
\begin{center}
\begin{tabular}{|c|c|c|c|c|c|}
\hline
\multirow{2}{*}{Methods} &Collision & Collision& Collision&Collision & Non- \\
 & in 1s & in 2s & in 3s & in 5s& collision\\

\hline
 STDAN   & 4.11 & 3.50 & 2.81  &  2.22 & 2.01\\
  STRAP-B  & 3.61 & 3.24& 2.67 &  2.18 & \textbf{1.87}\\
  STRAP-R  & \textbf{3.53} & \textbf{3.14} & \textbf{2.63}  & \textbf{2.16} & 1.87\\
\hline

\end{tabular}
\label{tab2}
\end{center}
\end{table}


\subsection{Ablation Study}
Table~\ref{tab3} presents an ablation study of STRAP by removing each key module. STRAP(–SE) excludes the spatial encoder, STRAP(–TE) excludes the temporal encoder, and STRAP(–RFF) excludes the risk feature fusion module. All models are trained with the risk-scaled loss on the NGSIM dataset, and all ablated models exhibit a drop in prediction accuracy compared to the comprehensive model, STRAP-R. The largest drop is observed in STRAP(–SE), which shows an increase in average RMSE of 14.6\% compared to STRAP-R, demonstrating that spatial interactions with surrounding vehicles are crucial for accurate long-term trajectory prediction. Removing the temporal encoder and the risk feature fusion module also increases the average RMSE by 3.9\% and 3.4\%, respectively, for STRAP(-TE) and STRAP(-RFF), confirming that each component contributes to the improvement of overall performance.

\begin{table}[htbp]
\caption{Prediction RMSE (m) in NGSIM on ablation models}
\begin{center}
\begin{tabular}{|c|c|c|c|c|c|c|}
\hline
Model & 1s & 2s & 3s & 4s & 5s & Average \\
\hline
STRAP(-SE) & 0.47   & 1.12 & 1.86 & 2.79 & 3.94 & 2.04\\
STRAP(-TE) & 0.45  & 1.04 & 1.68 & 2.50 & 3.56 & 1.85\\
STRAP(-RFF) & 0.44  & 1.01 & 1.68 & 2.51 &  3.56 & 1.84\\
STRAP-R & \textbf{0.37}  & \textbf{0.94} & \textbf{1.61} &  \textbf{2.47} & \textbf{3.52} & \textbf{1.78}\\
 \hline
\end{tabular}
\label{tab3}
\end{center}
\end{table}

\section{Conclusion}
In this paper, we propose STRAP, a spatial-temporal risk-attentive trajectory prediction framework that integrates a risk potential field to capture the perceived dangers from surrounding vehicles and generate more accurate trajectories that represent human driving behaviors. The framework consists of a spatial-temporal encoder and a risk-attentive feature fusion decoder, allowing risk cues to be embedded into the learned spatial-temporal features. Furthermore, a risk-scaled loss function is developed to specifically enhance prediction performance in high-risk scenarios. Experimental results on the NGSIM and HighD datasets show that the proposed method outperforms existing state-of-the-art models, and is able to produce more accurate trajectory predictions under high-risk scenarios.

In future work, we plan to extend our framework to more complex urban settings with richer road constraints beyond highways. We will also consider interactions among a wider range of agents, including vehicles, pedestrians, and cyclists, and develop corresponding risk field models to support more accurate multi-agent trajectory prediction.

\vspace{12pt}

\end{document}